\documentclass[journal,10pt]{IEEEtran}
\ifCLASSINFOpdf
\else
\fi
\usepackage{comment} 

\usepackage{amsmath,graphicx}

\usepackage{pstricks,graphicx}
\usepackage{setspace}
\usepackage{psfrag}
\usepackage{amssymb}
\usepackage{amsmath}
\usepackage{amsthm}

\newtheorem{remark}{Remark}

\begin{document}



\title{Training Deep Neural Networks \\ via Optimization Over Graphs}

\graphicspath{{figures/}}
%

\author{Guoqiang~Zhang and W.~Bastiaan Kleijn
\thanks{G.~Zhang is with the center of Audio, Acoustics and Vibration (CAAV), School of Computing and Communications, University of Technology, Sydney, Australia. Email: {guoqiang.zhang@uts.edu.au}}

\thanks{W.~B.~Kleijn is with is the school of Engineering and Computer Science, Victoria University of Wellington, New Zealand.
Email: {bastiaan.kleijn@ecs.vuw.ac.nz}}
}

\maketitle

\begin{abstract}
In this work, we propose to train a deep neural network by distributed optimization over a graph. Two nonlinear functions are considered: the rectified linear unit (ReLU) and a linear unit with both lower and upper cutoffs (DCutLU). The problem reformulation over a graph is realized by explicitly representing ReLU or DCutLU using a set of slack variables. We then apply the alternating direction method of multipliers (ADMM) to update the weights of the network layerwise by solving subproblems of the reformulated problem.  Empirical results suggest that the ADMM-based method is less sensitive to overfitting than the stochastic gradient descent  (SGD) and Adam methods. 
\end{abstract}



%
\IEEEpeerreviewmaketitle

\vspace{-2mm}
\section{Introduction}
\vspace{-1mm}
In the last decade, research on deep learning has made remarkable progress both in theoretical understanding and in practical applications (see \cite{Lecun15nature} for an overview).  Deep learning interprets data at multiple levels of abstraction, realized in a computational model with multiple processing layers. Each layer is composed of a set of simple nonlinear processing units (referred to as \emph{neurons}), which aims to transform the input into progressively more abstract representations \cite{Zeiler13, Yosinski15}. With the composition of multiple processing layers, the model is able to produce data representations that are required by various applications. 


In the literature, different types of deep neural networks (DNNs) have been proposed and applied in different applications. For instance, feedforward neural networks have been successfully applied in speech recognition \cite{Mohamed12,Dahl12}.  Convolutional neural networks  (CNNs) are popular in computer vision \cite{Krizhevsky12,Taigman14faceRecognition}. Recurrent neural networks (RNNs) have proven to be effective for mapping sequential inputs and outputs \cite{Graves13RNN,Sutskever14}.  

The common procedure for training a deep neural network is to iteratively adjust its parameters (referred to as \emph{weights})  such that the network approximates the input-output relations with increasing accuracy, referred to as  \emph{supervised learning}. 


The traditional supervised learning approach treats a neural network as a large complex model \cite{Lecun15nature} rather than decomposing it as a combination of many small nonlinear models. The standard procedure,  \emph{stochastic gradient descent (SGD)},  is to back-propagate gradients from the top layer down to the bottom layer on a mini-batch and then adjusts the weights accordingly. In recent years, various advanced methods have been proposed to use the gradient information smartly for either fast convergence  or automatic parameter adjustment, such as Adam \cite{Kingma17},  AdaGrad \cite{Duchi11AdaGrad} and RMSprop \cite{Hinton14RMSprop}.

In recent years, a new supervised learning paradigm has been proposed that decomposes a neural network as a combination of many small nonlinear models. In \cite{Carreira12}, the authors firstly proposed to decouple the nested structure of DNNs by introducing a set of auxiliary variables and a set of equality constraints. However, computation of the gradient is still required in their work to tackle the nonlinear functions of the neurons. The work of  \cite{Taylor16DNNADMM} avoids the gradient computation of  \cite{Carreira12} by using the alternating direction method of multipliers (ADMM) \cite{Boyd11ADMM}. However, \cite{Taylor16DNNADMM} needs to perform a computation at each and every neuron to be able to characterize its nonlinear operation. The Bregman iteration is used in \cite{Taylor16DNNADMM}  to produce stable algorithmic convergence. 

In this paper, we propose to train a deep neural network by reformulating the problem as an optimization over a factor graph $\mathcal{G}=(\mathcal{V},\mathcal{C})$ \cite{Sontag11ML,Meshi11ADMM}. Every node $r\in \mathcal{V}$ carries a convex function of its node variable while every factor  $c\in \mathcal{C}$ carries a nonlinear equality constraint in terms of the node variables connected to the factor. Our graphic formulation is able to handle rectified linear units  (ReLUs) (see \cite{Glorot11ReLU, Nair10ReLU}) and linear units with both upper and lower cutoffs (DCutLUs) at the layer-level. In particular, the ReLUs or DCutLUs are represented in terms of a set of slack variables, which lead to the equality constraints in the factor graph.   

We apply ADMM to solve the graph-based problem. Differently from \cite{Taylor16DNNADMM} which has to perform computations at the neuron-level, our proposed method is able to perform computations at the layer-level like the SGD and Adam. Experimental results on the MNIST dataset demonstrate that the new training method is less sensitive to overfitting than the SGD and Adam methods. Further, the performance of the new method on the test data is better than the SGD and Adam, which may be due to the flexibility of ADMM.

  
  
\vspace{-1mm}
\section{On Training a Deep Neural Network}
\vspace{-1mm}
Suppose we have a sequence of $m$ training samples, represented by an input matrix $D\in \mathbb{R}^{m\times n_{{in}}}$ and an output matrix $O\in \mathbb{R}^{m\times n_{out}}$, where the $q$'th row-vectors of $D$ and $O$ form an input-output pair.  Given $(D,O)$, we consider training a deep neural network with the weights $\{(W_i,b_i)|i=1,\ldots,N\}$ of $N$ layers, where for each $i$, $W_i\in\mathbb{R}^{n_{i-1}\times n_{i}}$ is a weight matrix and $b_i\in\mathbb{R}^{1\times n_{i}}$ a bias vector. To match the network with the training samples, we let $n_0=n_{in}$ and $n_{N}=n_{out}$. The objective is to find the proper weights $\{(W_i,b_i)\}$ so that the  network maps the input $D$ to the output $O$ as accurately as possible. 

Let us now define the operation of the individual layers. We use $V_i$ to denote the output of layer $i$, $i\leq N$.  We let $e$ be a (column) vector of ones.  $V_i$, $i\leq N-1$, is obtained by performing (element-wise) nonlinear operation on the matrix product $V_{i-1}W_i+eb_i$, denoted as $V_i = h_i(V_{i-1}W_i+eb_i)$. The popular forms for the nonlinear function $h_i$ are sigmoid, tanh and ReLU  \cite{Lecun15nature}. It is found in \cite{Nair10ReLU} that ReLU leads to fast convergence using SGD as compared to sigmoid and tanh. We consider ReLU and DCutLU in this paper. Formally, we define $h_i$ in the form of DCutLU as
\begin{align}
V_i=\min(\max(V_{i-1}W_i+eb_i ,l),u)\quad i\leq N-1,\label{equ:NN_cond0}
\end{align} where the $\max$ and $\min$ operators are element-wise, and $l$ and $u$ are the lower and upper threshold, respectively. ReLU is a special case of DCutLU by letting $(l,u)=(0,\infty)$. 

The procedure of training the above neural network can be formulated as 
\begin{align}
&\min_{\{V_i, W_i,b_i\}} \left[f_{N}(V_N;O)+\sum_{i=1}^{N} g_i(W_i,b_i)\right], \label{equ:NN}
\end{align}
where $f_N$ measures the difference between the output $V_N$ and the ground truth $O$,  $g_i$ is a penalty function on $(W_i,b_i)$, and $\{V_i, W_i,b_i\}$ satisfies (\ref{equ:NN_cond0}) and 
\begin{align}
& \quad   \;\; V_N=V_{N-1}W_N+e b_N. \label{equ:NN_cond1}
\end{align}

\vspace{-0mm}

\vspace{-3mm}
\section{Problem Reformulation Onto a Graph}
\vspace{-0mm}
In this section, we reformulate (\ref{equ:NN})-(\ref{equ:NN_cond1}) as an optimization over a factor graph. We first represent the nonlinear function (\ref{equ:NN_cond0}) by introducing a set of slack variables.  Specifically,  (\ref{equ:NN_cond0}) can be rewritten as 
\begin{align}
X_i&=V_{i-1}W_i+e b_i \label{equ:NN_cond2} \\
X_i+Y_i&=\max(V_{i-1}W_i+e b_i, l) \label{equ:NN_cond3} \\
X_i+Y_i+Z_i&=V_i=\min(X_i+Y_i,u), \label{equ:NN_cond4}
\end{align}
where for each $i\in \{1,\ldots, N-1\}$, we introduced three slack matrices $X_i$, $Y_i$ and $Z_i$ to characterize the effect of the upper and lower cutoffs of the function at $u$ and $l$. 

Next, we argue that the $\min$ and $\max$ operators in (\ref{equ:NN_cond3})-(\ref{equ:NN_cond4}) can be expressed in terms of constraints on $(X_i,Y_i,Z_i)$. To do so, we introduce two index sets for each layer $i$: 
\begin{align}
\Omega_i^l &= \{ (q,j) | x_{i,qj} <l\}   \\
\Omega_i^u &= \{ (q,j) | x_{i,qj} > u\},  
\end{align} 
\hspace{-1.2mm} where $x_{i,qj}$ is the $(q,j)$ element of $X_i$. At the moment, one can think of $\Omega_i^l$ and $\Omega_i^u$ as two sets that are preset already, imposing constraints on $X_i$. We will explain later how to update $(\Omega_i^l,\Omega_i^u)$ iteratively. Given a set $\Omega$,  we let $\mathcal{P}_\Omega(X)$ denote the subset of the elements of $X$ specified by the indices of $\Omega$. The $\max$ operator in (\ref{equ:NN_cond3}) can be characterized as 
 \vspace{-2mm}
\begin{subequations}
\begin{alignat}{1}
\mathcal{P}_{\Omega_i^l}(X_i)&< l \\
\mathcal{P}_{\Omega_i^l}(X_i)+\mathcal{P}_{\Omega_i^l}(Y_i)&=l \\  
\mathcal{P}_{\bar{\Omega}_i^l}(X_i)&\geq l \\
\mathcal{P}_{\bar{\Omega}_i^l}(Y_i) &= 0,
\end{alignat}
\label{equ:Y_cond}  
\end{subequations}
\hspace{-1.4mm}where $\bar{\Omega}$ denotes the complement of  $\Omega$. By inspection of (\ref{equ:NN_cond3}) and (\ref{equ:NN_cond4}), we conclude that $Y_i$ and $Z_i$ are decoupled given $X_i$.  The $\min$ operator in (\ref{equ:NN_cond4}) can thus be characterized as  
\vspace{-2mm}
\begin{subequations}
\begin{alignat}{1}
\mathcal{P}_{\Omega_i^u}(X_i)&> u \\ 
\mathcal{P}_{\Omega_i^u}(X_i)+\mathcal{P}_{\Omega_i^u}(Z_i)&=u \\
\mathcal{P}_{\bar{\Omega}_i^u}(X_i)&\leq u \\
\mathcal{P}_{\bar{\Omega}_i^u}(Z_i)&=0.
\end{alignat} 
\label{equ:Z_cond} 
\end{subequations}
\hspace{-1.4mm}To briefly summarize, we use the constraints (\ref{equ:Y_cond}) and (\ref{equ:Z_cond}) to replace the $\min$ and $\max$ operations in (\ref{equ:NN_cond3})-(\ref{equ:NN_cond4}).

\begin{figure}[t]
\centering
\begin{footnotesize}
  \includegraphics[width=80mm]{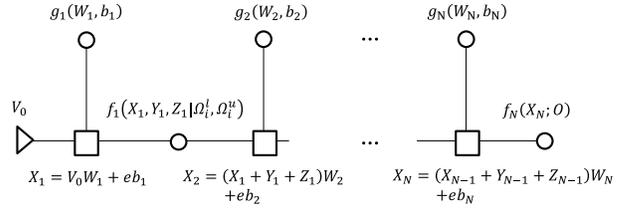}
\end{footnotesize}
\caption{\small  Problem reformulation over a factor graph $\mathcal{G}=(\mathcal{V},\mathcal{C})$. $\circ$ is a node in $\mathcal{V}$ and {\small $\square$} is a  factor in $\mathcal{C}$. $\rhd$ represents constant inputs to the graph, where  $X_0=V_0=D$ is the data input.}   \vspace{-3mm} \label{fig:graph_NN}
\vspace{-2mm}
\end{figure}

\begin{sloppypar}
Based on the above analysis, the training problem (\ref{equ:NN})-(\ref{equ:NN_cond1}) can be reformulated as  \vspace{-1mm}
\begin{align}
&\hspace{-8mm}\min_{\hspace{5mm}\tiny\begin{array}{ll} \hspace{-1mm}\{W_i\hspace{-0.2mm},\hspace{-0.3mm}b_i \hspace{-0.2mm} \\ \hspace{-0.2mm}X_i\hspace{-0.2mm},\hspace{-0.3mm}Y_i\hspace{-0.2mm}, \hspace{-0.3mm} Z_i \hspace{-0.2mm}\} \end{array}}\hspace{-7.mm} f_N(\hspace{-0.2mm}X_{N}\hspace{-0.2mm};\hspace{-0.3mm}O\hspace{-0.2mm}) \hspace{-0.6mm}+\hspace{-1mm}  \sum_{i=1}^N \hspace{-0.3mm}g_i(W_i \hspace{-0.2mm},\hspace{-0.3mm}b_i)\hspace{-0.6mm} +\hspace{-1mm}\sum_{i= 1}^{N-1} \hspace{-0.9mm}f_i(X_i\hspace{-0.2mm},\hspace{-0.3mm}Y_i\hspace{-0.2mm},\hspace{-0.3mm}Z_i\hspace{-0.2mm}| \hspace{-0.2mm}\Omega_i^l\hspace{-0.2mm},\hspace{-0.3mm}\Omega_i^u)\hspace{-2mm} \label{equ:G_pro}\\
&\textrm{s. t. } X_{i}\hspace{-0.6mm}=\hspace{-0.6mm}(X_{i-1}\hspace{-0.6mm}+\hspace{-0.6mm}Y_{i-1}\hspace{-0.6mm}+\hspace{-0.6mm}Z_{i-1})W_i\hspace{-0.6mm}+\hspace{-0.6mm} e b_i \hspace{2mm} \forall i\hspace{-0.6mm} =\hspace{-0.6mm}1,\ldots,\hspace{-0.3mm}N, \label{equ:G_cond} 
\end{align}
where  $(X_0,\hspace{-0.1mm}Y_0,\hspace{-0.1mm}Z_0)=(V_0,\hspace{-0.1mm}0,\hspace{-0.1mm}0)$ and each $f_i(X_i,\hspace{-0.1mm}Y_i,\hspace{-0.1mm}Z_i|\Omega_i^l,\hspace{-0.1mm}\Omega_i^u)$ can be taken as a summation of indicator functions, each defined by one constraint in (\ref{equ:Y_cond})-(\ref{equ:Z_cond}), given by \vspace{-0.5mm}
\begin{align}
f_i(X_i,&Y_i,Z_i|\Omega_i^l,\Omega_i^u) \hspace{-0.7mm}=\hspace{-1.1mm} \Big[1_{\mathcal{P}_{\Omega_i^l}(X_i)< l} \hspace{-0.7mm}+\hspace{-0.6mm}1_{\mathcal{P}_{\bar{\Omega}_i^l}(Y_i) = 0} \hspace{-0.6mm} + \hspace{-0.6mm}1_{\mathcal{P}_{\bar{\Omega}_i^l}(X_i)\geq l} \nonumber \\ 
&+1_{\mathcal{P}_{\Omega_i^l}(X_i)+\mathcal{P}_{\Omega_i^l}(Y_i)=l}+1_{\mathcal{P}_{\Omega_i^u}(X_i)> u} +1_{\mathcal{P}_{\bar{\Omega}_i^u}(Z_i)=0} \nonumber\\ 
&+1_{\mathcal{P}_{\bar{\Omega}_i^u}(X_i)\leq u} +1_{\mathcal{P}_{\Omega_i^u}(X_i)+\mathcal{P}_{\Omega_i^u}(Z_i)=u}\Big], 
\label{equ:G_fi}\end{align}
where  the indicator function $1_{(\cdot)}$ equals to 0 when its constraint is satisfied and equals to $+\infty$ otherwise.  
\end{sloppypar}

Eqn. (\ref{equ:G_pro})-(\ref{equ:G_fi}) define a problem over a factor graph $\mathcal{G}=(\mathcal{V},\mathcal{C})$ (see  \cite{Meshi11ADMM,Sontag11ML,Wainwright08Overview}), where every node $r\in \mathcal{V}$ carries a (convex) component function of (\ref{equ:G_pro}) and  every factor $c\in \mathcal{C}$ carries an (nonlinear) equality constraint of (\ref{equ:G_cond}) (see Fig. \ref{fig:graph_NN} for demonstration).

\begin{remark}
If the ReLU is chosen for layer $i$ of the network, one can simply ignore $Z_i$ and $\Omega_i^u$ and let $l=0$ in (\ref{equ:G_pro})-(\ref{equ:G_fi}). 
\vspace{-2mm}
\end{remark}

\vspace{-2mm}
\section{Distributed Optimization Over a Graph}
\vspace{-1mm}
\label{sec:optimization}

We note that (\ref{equ:G_pro})-(\ref{equ:G_fi}) is a nonconvex optimization because of the nonlinear equality constraints (\ref{equ:G_cond}).  We solve (\ref{equ:G_pro})-(\ref{equ:G_fi}) in an iterative fashion using ADMM by solving convex subproblems. It is worth noting that ADMM has already been successfully applied for solving nonnegative matrix factorization (NMF) \cite{Hajinezhad16NMF}, which is nonconvex. 
 
\vspace{-2mm}
\subsection{Augmented Lagrangian function}
\vspace{-1mm}
To apply ADMM,  we introduce a Lagrange multiplier $\Lambda_i$ for the $i$th equality constraint in (\ref{equ:G_cond}). We build an augmented Lagrangian function as
\begin{align}
&\hspace{-2mm} L_{\{\rho_i\}}(\{X_i,Y_i,Z_i,b_i,W_i,\Lambda_{i}\},X_N | \{\Omega_i^l,\Omega_i^u\}) \nonumber \\
&\hspace{-2mm} =\hspace{-0.3mm}f_N(\hspace{-0.2mm}X_{N}\hspace{-0.2mm};\hspace{-0.3mm}O\hspace{-0.2mm}) \hspace{-0.6mm}+\hspace{-1mm}  \sum_{i=1}^N \hspace{-0.3mm}g_i(W_i \hspace{-0.2mm},\hspace{-0.3mm}b_i)\hspace{-0.6mm} +\hspace{-1mm}\sum_{i= 1}^{N-1} \hspace{-0.7mm}f_i(X_i\hspace{-0.2mm},\hspace{-0.3mm}Y_i\hspace{-0.2mm},\hspace{-0.3mm}Z_i\hspace{-0.2mm}|\hspace{-0.3mm}\Omega_i^l\hspace{-0.2mm},\hspace{-0.3mm}\Omega_i^u)\hspace{-0.3mm}\nonumber \\
 &+\hspace{-1mm}\sum_{i=1}^N p_{i,\rho_i}((X_{i-1}\hspace{-0.2mm},\hspace{-0.3mm}Y_{i-1}\hspace{-0.2mm},\hspace{-0.3mm}Z_{i-1}\hspace{-0.2mm}),\hspace{-0.3mm}X_i\hspace{-0.2mm},(\hspace{-0.3mm}W_i\hspace{-0.2mm},\hspace{-0.3mm}b_i\hspace{-0.2mm}),\hspace{-0.3mm}\Lambda_i \hspace{-0.2mm}), \label{equ:augLag}
\end{align}
where for each $i=1,\ldots, N$, $p_{i,\rho_i}(\cdots)$ is defined as  
\begin{align}
\hspace{-3mm}p_{i,\rho_i}(\cdots)& \hspace{0mm}=\Big[\frac{\rho_{i}}{2}\| X_{i}\hspace{-0.5mm}-\hspace{-0.5mm}(X_{i-1}\hspace{-0.5mm}+\hspace{-0.5mm}Y_{i-1}\hspace{-0.5mm}+\hspace{-0.5mm}Z_{i-1})W_i\hspace{-0.5mm}-\hspace{-0.5mm} e b_i \|^2 \nonumber \\
&  \hspace{3.5mm}+\hspace{-0.5mm}\langle \Lambda_{i}, X_{i}\hspace{-0.5mm}-\hspace{-0.5mm}(X_{i-1}\hspace{-0.5mm}+\hspace{-0.5mm}Y_{i-1}\hspace{-0.5mm}+\hspace{-0.5mm}Z_{i-1})W_i \hspace{-0.5mm}-\hspace{-0.5mm} e b_i  \rangle  \Big], \label{equ:qi}
\end{align}
where $\rho_i>0$, $(X_{0},Y_{0},Z_{0})=(V_0,0,0)$, and $\langle \cdot, \cdot \rangle$ denotes dot product. We note that differently from the single learning rate of SGD, each layer $i$ possesses a positive parameter $\rho_i$, which can be treated as a layer-oriented learning rate.  

Our objective now is to reach a saddle point of the Lagrangian function $L_{\{\rho_i\}}$ by  minimizing over $\{X_i,Y_i,Z_i,b_i,W_i,\}\cup X_N$ and maximizing over $\{\Lambda_{i}\}$. A saddle point would satisfy the equality constraints (\ref{equ:G_cond}).

\begin{table}[t!]
\caption{\small ADMM updating procedure}
\centering
\begin{tabular}{l}
\hline
{\small \textrm{Initialize}: $\{\hat{W}_i, \hat{b}_i,|i=1,\ldots, N\}$ } \\
{\small Repeat }\\
   \hspace{1mm} {\small Feed $X_0$ to the DNN and initialize $\{\hat{X}_i,\hat{Y}_i,\hat{Z}_i, \hat{\Omega}_i^l,\hat{\Omega}_i^u\}$} \\	
   \hspace{1mm} Let $\{\hat{\Lambda}_i=0| i=1,\ldots, N\} $ \\
   \hspace{1mm} {\small For $i=N,N-1,\ldots, 1$ do} \\
   \hspace{3mm} $(\hat{X}_i^{new},\hat{Y}_i^{new},\hat{Z}_i^{new})$\\
   \hspace{3mm} $=\arg\min L_{i}((X_i,Y_i,Z_i), (\hat{W}_i,\hat{b}_i),\hat{\Lambda}_i|\Omega_i^u,\Omega_i^l)$ \\
       \hspace{3mm} {\small $\hat{\Lambda}_i^{new}\hspace{-0.5mm}=\hspace{-0.5mm}\rho_i(\hat{X}_{i}^{new}\hspace{-0.5mm}-\hspace{-0.5mm}(\hat{X}_{i-1}\hspace{-0.5mm}+\hspace{-0.5mm}\hat{Y}_{i-1}\hspace{-0.5mm}+\hspace{-0.5mm}\hat{Z}_{i-1})\hat{W}_i\hspace{-0.5mm}-\hspace{-0.5mm} e\hat{b}_i)$ }\\
        \hspace{3mm} {\small $(\hat{W}_i^{new},\hat{b}_i^{new})$} \\
       \hspace{4.1mm}{\small $ =\arg\min L_{i}((\hat{X}_i^{new},\hat{Y}_i^{new},\hat{Z}_i^{new}), (W_i,b_i),\hat{\Lambda}^{new}_i|\Omega_i^u,\Omega_i^l)$ }    \\        
   \hspace{1mm}  {\small End for} \\
   \hspace{2mm}$(\hat{W}_i,\hat{b}_i)=(\hat{W}_i^{new},\hat{b}_i^{new})$ for all $i$\\
{\small Until some stopping criterion is met} \\
\hline
\end{tabular}
\label{table:NN_ADMM}
\vspace{-5.5mm}
\end{table}

\vspace{-2mm}
\subsection{Blockwise parameter updating using ADMM}
\vspace{-1mm}
We now consider optimizing the Lagrangian function $L_{\{\rho_i\}}$. We follow a similar updating procedure as the SGD and Adam methods \cite{Kingma17}. That is, at each iteration, we initialize all the variables and index sets of  $L_{\{\rho_i\}}$ by feeding $D$ to the network through the forward computation. We then update all the variables of $L_{\{\rho_i\}}$ blockwise through backward computation. Differently from SGD which computes gradient directly, the variables of $L_{\{\rho_i\}}$ are updated by solving small-size optimization problems.

Suppose we finished updating variables of layer $i+1$ and would like to update $(X_i,Y_i,Z_i), (W_i,b_i)$ and $\Lambda_i$ of layer $i$.  We first simplify $L_{\{\rho_i\}}$ by removing irrelevant components,
\begin{align}
&\hspace{-1mm}L_{i}((X_i,Y_i,Z_i), (W_i,b_i),\Lambda_i|\Omega_i^u,\Omega_i^l) \nonumber \\
&\hspace{-1mm} = p_{i+1,\rho_{i+1}}((X_{i}\hspace{-0.2mm},\hspace{-0.3mm}Y_{i}\hspace{-0.2mm},\hspace{-0.3mm}Z_{i}\hspace{-0.2mm}),\hspace{-0.3mm}\hat{X}_{i+1}^{new}\hspace{-0.2mm},(\hspace{-0.3mm}\hat{W}_{i+1}\hspace{-0.2mm},\hspace{-0.3mm}\hat{b}_{i+1})\hspace{-0.2mm},\hspace{-0.3mm}\hat{\Lambda}_{i+1}^{new} \hspace{-0.2mm}) \nonumber \\
& \;\;\;+p_{i,\rho_i}((\hat{X}_{i-1}\hspace{-0.2mm},\hspace{-0.3mm}\hat{Y}_{i-1}\hspace{-0.2mm},\hspace{-0.3mm}\hat{Z}_{i-1}\hspace{-0.2mm}),\hspace{-0.3mm}X_i\hspace{-0.2mm},(\hspace{-0.3mm}W_i\hspace{-0.2mm},\hspace{-0.3mm}b_i\hspace{-0.2mm}),\hspace{-0.3mm}\Lambda_i \hspace{-0.2mm})+g_i(W_i,b_i) \nonumber
\end{align}
\begin{align}
&\hspace{-25mm}+ f_i(X_i\hspace{-0.2mm},\hspace{-0.3mm}Y_i\hspace{-0.2mm},\hspace{-0.3mm}Z_i\hspace{-0.2mm}|\hspace{-0.3mm}\Omega_i^l\hspace{-0.2mm},\hspace{-0.3mm}\Omega_i^u),\label{equ:Li}
\end{align}
where $\hat{X}_{i+1}^{new}$ and $\hat{\Lambda}_{i+1}^{new}$ are the new estimate obtained from the computation at layer $i+1$. 
By following the ADMM updating procedure, we first compute $(\hat{X}_i^{new},\hat{Y}_i^{new},\hat{Z}_i^{new})$ by optimizing $L_{i}$ with $(\hat{W}_i,\hat{b}_i)$ and $\hat{\Lambda}_i$ fixed. We then compute $\hat{\Lambda}_i^{new}$ using $\hat{X}_i^{new}$ through dual ascent. Finally, we compute $(\hat{W}_i^{new}$,$\hat{b}_i^{new})$  by optimizing $L_i$ with $(\hat{X}_i^{new},\hat{Y}_i^{new},\hat{Z}_i^{new})$ and $\hat{\Lambda}_i^{new}$ fixed. See Table~\ref{table:NN_ADMM} for the updating procedure. 

For the top layer $i=N$, the function $L_N$ takes the form: 
\begin{align}
&L_{N}(X_N, (W_N,b_N),\Lambda_N) = f_N(X_N;O)+g_N(W_N,b_N) \nonumber\\
&\hspace{12mm}+p_{N,\rho_N}((\hat{X}_{N-1}\hspace{-0.2mm},\hspace{-0.3mm}\hat{Y}_{N-1}\hspace{-0.2mm},\hspace{-0.3mm}\hat{Z}_{N-1}\hspace{-0.2mm}),\hspace{-0.3mm}X_N\hspace{-0.2mm},(\hspace{-0.3mm}W_N\hspace{-0.2mm},\hspace{-0.3mm}b_N\hspace{-0.2mm}),\hspace{-0.3mm}\Lambda_N \hspace{-0.2mm}), \nonumber
\end{align}
where there is no $Y_N$ and $Z_N$. For this layer, only $\hat{X}_N^{new}$ is computed by optimizing $L_N$ in Table~\ref{table:NN_ADMM}. 

\begin{remark}
In (\ref{equ:Li}),  $(\hat{W}_{i+1}, \hat{b}_{i+1})$ is used instead of  $(\hat{W}_{i+1}^{new}, \hat{b}_{i+1}^{new})$, which is found to be much more stable through experiments.  \vspace{-3.5mm}
\end{remark}

\begin{table}[t!]
\caption{ Computing $(\hat{X}_i^{new},\hat{Y}_i^{new},\hat{Z}_i^{new})$ for each $i < N$ in Table~\ref{table:NN_ADMM}.}\
\vspace{-2mm}
\centering
\begin{tabular}{l}
\hline
{\small \textrm{Let}: $(\hat{X}_{i}^c,\hat{Y}_{i}^c,\hat{Z}_{i}^c)\hspace{-0.6mm}=\hspace{-0.6mm}(\hat{X}_i,\hat{Y}_i,\hat{Z}_i)$ and $(\hat{\Gamma}_i^x,\hat{\Gamma}_i^y,\hat{\Gamma}_i^z)\hspace{-0.6mm}=\hspace{-0.6mm}(0,0,0) $} \\
\hspace{0mm} {\small $(\hat{X}_i^{new},\hat{Y}_i^{new},\hat{Z}_i^{new})$} \\
\hspace{0mm} {\small $=\hspace{-0.5mm}\arg\min\hspace{-0.5mm} L_{i,\beta_i} ((X_i,Y_i,Z_i),(\hat{\Gamma}_i^x,\hat{\Gamma}_i^y,\hat{\Gamma}_i^z) ,(\hat{X}_{i}^c,\hat{Y}_{i}^c,\hat{Z}_{i}^c))  $} \\
 \hspace{0mm} {\small $\hat{\Gamma}_i^{x,new} = \beta_i(\hat{X}_i^{new}-\hat{X}_{i}^c)$} \\
 \hspace{0mm} {\small $\hat{\Gamma}_i^{y,new} = \beta_i(\hat{Y}_i^{new}-\hat{Y}_{i}^c) $} \\
 \hspace{0mm} {\small $\hat{\Gamma}_i^{z,new} = \beta_i(\hat{Z}_i^{new}-\hat{Z}_{i}^c)$} \\           
 \hspace{0mm} {\small $(\hat{X}_{i}^{c,new},\hat{Y}_{i}^{c,new},\hat{Z}_{i}^{c,new})=\hspace{-0.5mm}\arg\min\hspace{-0.5mm}L_{i,\beta_i} ((\hat{X}_i^{new},\hat{Y}_i^{new},$} \\
 \hspace{15mm} {\small $\hat{Z}_i^{new}),(\hat{\Gamma}_i^{x,new},\hat{\Gamma}_i^{y,new},\hat{\Gamma}_i^{z,new}) ,(X_{i}^c,Y_{i}^c,Z_{i}^c)) $} \\	
   $(\hat{X}_{i}^{new},\hat{Y}_{i}^{new},\hat{Z}_{i}^{new})=(\hat{X}_{i}^{c,new},\hat{Y}_{i}^{c,new},\hat{Z}_{i}^{c,new})$ \\
\hline 
\end{tabular}
\label{table:update_XYZ}
\vspace{-2mm}
\end{table}

\vspace{-2mm}
\subsection{Handling of the indicator function}
\vspace{-1mm}

The function $f_i(\cdot)$ in (\ref{equ:Li}) is composed of a set of indicator functions, which makes it difficult to compute $(\hat{X}_i^{new},\hat{Y}_i^{new},\hat{Z}_i^{new})$ in Table~\ref{table:NN_ADMM}.  To facilitate the computation,  we introduce the auxiliary variables $(X_{i}^c,Y_{i}^c,Z_{i}^c)$ to replace $(X_i, Y_i, Z_i)$ in $f_i(X_i,Y_i,Z_i | \Omega_i^l,\Omega_i^u)$ with the constraints $X_{i}^c=X_i^c$, $Y_{i}^c=Y_i$ and $Z_{i}^c=Z_i$. We then apply ADMM again to handle the three equality constraints. To do so, we build a new augmented Lagrangian as
\begin{align}
&L_{i,\beta_i} ((X_i,Y_i,Z_i),(\Gamma_i^x,\Gamma_i^y,\Gamma_i^z) ,(X_{i}^c,Y_{i}^c,Z_{i}^c)) \nonumber \\
& = p_{i+1,\rho_{i+1}}((X_{i}\hspace{-0.2mm},\hspace{-0.3mm}Y_{i}\hspace{-0.2mm},\hspace{-0.3mm}Z_{i}\hspace{-0.2mm}),\hspace{-0.3mm}\hat{X}_{i+1}^{new}\hspace{-0.2mm},(\hspace{-0.3mm}\hat{W}_{i+1}\hspace{-0.2mm},\hspace{-0.3mm}\hat{b}_{i+1})\hspace{-0.2mm},\hspace{-0.3mm}\hat{\Lambda}_{i+1}^{new} \hspace{-0.2mm}) \nonumber \\
& \;\;\;+p_{i,\rho_i}((\hat{X}_{i-1}\hspace{-0.2mm},\hspace{-0.3mm}\hat{Y}_{i-1}\hspace{-0.2mm},\hspace{-0.3mm}\hat{Z}_{i-1}\hspace{-0.2mm}),\hspace{-0.3mm}X_i\hspace{-0.2mm},(\hspace{-0.3mm}W_i\hspace{-0.2mm},\hspace{-0.3mm}b_i\hspace{-0.2mm}),\hspace{-0.3mm}\Lambda_i \hspace{-0.2mm})+g_i(W_i,b_i) \nonumber\\
&+\hspace{-0.7mm}f_i(X_{i}^c\hspace{-0.2mm},\hspace{-0.3mm}Y_{i}^c\hspace{-0.2mm},\hspace{-0.3mm}Z_{i}^c\hspace{-0.2mm}|\hspace{-0.3mm}\Omega_i^l\hspace{-0.2mm},\hspace{-0.3mm}\Omega_i^u)\hspace{-0.7mm} +\hspace{-0.7mm}\frac{\beta_i}{2}\|X_i\hspace{-0.7mm}-\hspace{-0.7mm}X_{i}^c\|^2+\hspace{-0.7mm}\langle \Gamma_i^x, X_i\hspace{-0.9mm}-\hspace{-0.9mm}X_{i}^c\rangle\hspace{-0.8mm} \nonumber \\
 &\hspace{-1.5mm}+\hspace{-0.8mm}\frac{\beta_i}{2}\|Y_i\hspace{-0.8mm}-\hspace{-0.8mm}Y_{i}^c\|^2\hspace{-0.8mm}+\hspace{-0.8mm}\langle \Gamma_i^y, Y_i\hspace{-0.8mm}-\hspace{-0.8mm}Y_{i}^c\rangle \hspace{-0.8mm}+\hspace{-0.8mm}\frac{\beta_i}{2}\|Z_i\hspace{-0.8mm}-\hspace{-0.8mm}Z_{i}^c\|^2\hspace{-0.7mm} +\hspace{-0.6mm}\langle \Gamma_i^z, Z_i\hspace{-0.6mm}-\hspace{-0.6mm}Z_{i}^c\rangle, \nonumber
\end{align}
where $\{\Gamma_i^x,\Gamma_i^y,\Gamma_i^z\}$ are the Lagrange multipliers, and $\beta_i>0$ which has a similar role as $\rho_i$ in $L_{\{\rho_i\}}$. 
We update the three sets of variables  $({X}_i,{Y}_i,{Z}_i)$,  $(\Gamma_i^x,\Gamma_i^y,\Gamma_i^z)$ and $({X}_{i}^c,{Y}_{i}^c,{Z}_{i}^c)$ one after another (see Table~\ref{table:update_XYZ}). To reduce the computational time, we only update the above variables once instead of multiple iterations.    

To briefly summarize, at each iteration, the proposed algorithm performs both forward and backward computations. The forward computation initializes all variables and index sets while the backward computation updates all the variables and the network weights. The algorithm has a set of learning rates $\{\rho_i\}\cup \{\beta_i\}$, which provides great flexibility to fine-tune the algorithm to have fast convergence (See the first experiment of Section~\ref{sec:experiment} about the parameter setup). 

\vspace{-2mm}
\section{Experimental results}
\label{sec:experiment}
\vspace{-1mm}
In the simulation, we considered the handwritten-digit recognition problem by using MNIST with the standard division of the training (60000 samples) and test  (10000 samples) datasets. In doing so, we built a DNN of three layers ($N=3$), where the first and second hidden layer consists of 500 and 600 neurons, respectively. The output function was chosen as the summation of  the individual cross-entropy functions (\cite{CoverBook06}). The function $g_i(W_i,b_i)$ was chosen as $\frac{0.1}{2}\|(W_i,b_i)\|^2$. The mini-batch size was set as 3000.  The entire training dataset thus consisted of 20 minibatches.  

We note that the cross-entropy function makes it difficult to compute $\hat{X}_N^{new}$ analytically in Table~\ref{table:NN_ADMM}.  When updating the above variable at each iteration, we approximate each cross-entropy term by a quadratic function around the most recent estimate, where the quadratic coefficient is set to 0.05 and the linear coefficient is set to the gradient. 

We evaluated the proposed method (referred to \emph{ADMM}) with two proof-of-concept experiments. In the first experiment, we tested ADMM, SGD and Adam \cite{Kingma17} using only the ReLUs.  In the second experiment, we studied how the learning rates $\{\rho_i\}\cup\{\beta_i\}$ affect the convergence speed of ADMM for both ReLUs and DCutLUs.    

\subsubsection{Comparison with the state-of-the-art} In addition to ADMM, we also evaluated SGD and Adam \cite{Kingma17}, where Adam represents the state-of-the-art training method. The goal is to study the convergence properties of the proposed algorithm. The learning rate of SGD was chosen as 0.3 (producing stable and fast convergence among $\{0.1, 0.2, 0.3,0.4\}$).  Adam was implemented by following \cite{Kingma17} directly. When running SGD and Adam, the gradient of ReLU at zero is set to $0$. Finally the learning rates of ADMM were set as  $\rho_3=0.05$, $\rho_2=\beta_2=0.1$ and $\rho_1=\beta_1=0.2$. The basic principle is to set the learning rates $\rho_i$ and $\beta_i$ of layer $i$ slightly larger than $\rho_{i+1}$ and $\beta_{i+1}$ of layer $i+1$.  

The experimental results are displayed in Fig.~\ref{fig:perCompare} (a). It is seen that the performance gap of ADMM between the test data and training data  is relatively stable compared to that of Adam and SGD. Furthermore, ADMM performs better than Adam and SGD on the test data, where the recognition accuracy for the test data at the last iteration is: 98.41(ADMM), 98.23(Adam) and 97.98(SGD). The better performance of ADMM might be due to the introduction of layer-oriented learning rates $\{\rho_i,\beta_i\}$. 

The computational time of the three methods was measured on an Apple MacBook Pro and is summarized in Table~\ref{tab:time}. In general, ADMM is somewhat more expensive than SGD and Adam because it consumes more memory due to the auxiliary variables and involves solving a set of small-size optimization problems per min-batch.      
\begin{table}[h!]
\caption{\small Average execution times (per mini-batch) and their standard deviations for the four methods. } 
\label{tab:time}
\centering
\begin{tabular}{|c|c|c|c|c|}
\hline
&\hspace{-2mm} {{\scriptsize SGD (ReLU)}}\hspace{-1.8mm}
& \hspace{-1.8mm}{\scriptsize Adam (ReLU)}  \hspace{-1.8mm}
& \hspace{-1.8mm}{\scriptsize ADMM (ReLU)} \hspace{-1.8mm} 
& \hspace{-1.8mm}\scriptsize{ADMM (DCutLU)} \hspace{-1.8mm}  \\
\hline  
\hspace{-1.8mm}\scriptsize{ave. (second)} \hspace{-1.8mm} & \footnotesize{0.2257} & \footnotesize{0.2398}& \footnotesize{0.9373}& \footnotesize{1.446} \\ 
\hline
\hspace{-1.5mm}\scriptsize{std }  \hspace{-1.5mm} & \footnotesize{0.0444} & \footnotesize{0.0416} & \footnotesize{0.0851} & \footnotesize{0.0949} \\ 
\hline
\end{tabular}
\vspace{-3mm}
\end{table}

\subsubsection{Effect of different learning rates on convergence speed}


In this experiment, we studied how the learning rates $\{\rho_i,\beta_i\}$ affect the convergence speed of ADMM for both ReLUs and DCutLUs (where $(l,u)=(0,1)$). To simplifying the evaluation, we let all $\rho_i$ and $\beta_i$ to be the same per experiment. For each learning rate, we counted the number of iterations over entire training dataset until the average cross-entropy reaches 0.05.   

\begin{figure}[t!]
\centering
\includegraphics[width=85mm]{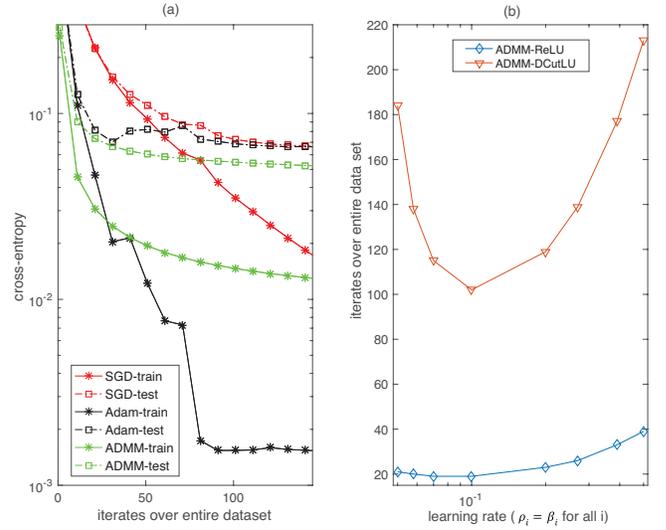}
\caption{\small  Performance comparison. Subplot~(a) displays the performance of SGD, Adam and ADMM using only ReLUs. Subplot (b) shows the number of iterations over entire training dataset needed to reach a threshold (0.05) of average cross-entropy for each learning rate of ADMM. }  \label{fig:perCompare} \vspace{-6mm}
\end{figure}

The convergence results are displayed in Fig.~\ref{fig:perCompare} (b). It is seen that the learning rate indeed affects the convergence speed. Further, it is observed that ReLU needs significantly fewer iterations than DCutLU. Table~\ref{tab:time} also shows that the computational time of ReLU is lower than that of DCutLU. This suggests that ReLU is a better choice in practice.  

\vspace{-0.6mm}
\vspace{-1mm}

\begin{remark}
At the moment, the convergence of the proposed method is only demonstrated by experiments. We leave the theoretical convergence analysis for future investigation. 
\end{remark}

\vspace{-2mm}
\section{Conclusions}
\vspace{-1mm}
We have proposed a new algorithm for training a DNN by performing optimization over a factor graph. The key step is to  explicitly represent the ReLUs or DCutLUs by a set of slack variables, which enables layer-level computation rather than neuron-level computation as in \cite{Taylor16DNNADMM}. Experimental results indicate that the new algorithm is less sensitive to over-fitting than two references.  One future research direction is to adjust the learning rates $\{\rho_i\}$ and $\{\beta_i\}$ of the new algorithm automatically, which likely will lead to good convergence speed for various learning problems.

\ifCLASSOPTIONcaptionsoff
  \newpage
\fi

\bibliographystyle{IEEEtran}
\bibliography{sigProcessing}

\end{document}